\documentclass[pmlr]{jmlr}

\usepackage{times}
\usepackage[utf8]{inputenc}
\usepackage{cite}
\usepackage{natbib}
\usepackage{hyperref}
\usepackage{url}
\usepackage{algorithm}
\usepackage{algorithmic} 
\usepackage{amsmath,bm}
\usepackage{bbm}
\usepackage{microtype}
\usepackage{graphicx}
\usepackage{parskip}
\usepackage{caption}
\usepackage{subcaption}
\usepackage{tikz}
\usepackage{color}
\usepackage{booktabs}
\usepackage{rotating}
\usepackage[percent]{overpic}
\usepackage{multicol}
\usepackage{float}
\usepackage{wrapfig}
\newcommand{\cut}[1]{}

\newcommand{\name}{MLP-GAL}

\newcommand{\keypoint}[1]{\vspace{0.0cm}\noindent\textbf{#1}\quad}
\title{Meta-Learning Transferable Active Learning Policies by Deep Reinforcement Learning}
\begin{document}




\author{\Name{Kunkun Pang} \Email{k.pang@ed.ac.uk}\\
  \addr Institute of Perception, Action and Behaviour, University of Edinburgh
  \AND
  \Name{Mingzhi Dong} \Email{mingzhi.dong.13@ucl.ac.uk}\\
  \addr Department of Statistics, University College London
  \AND
  \Name{Yang Wu} \Email{yangwu@rsc.naist.jp}\\
  \addr Nara Institute of Science and Technology
  \AND
  \Name{Timothy Hospedales} \Email{t.hospedales@ed.ac.uk}\\
  \addr Institute of Perception, Action and Behaviour, University of Edinburgh
 }
\maketitle
\begin{abstract}
    Active learning (AL) aims to enable training high performance classifiers with low annotation cost by predicting which subset of unlabelled instances would be most beneficial to label. The importance of AL has motivated extensive research, proposing a wide variety of manually designed AL algorithms with diverse theoretical and intuitive motivations.
    In contrast to this body of research, we propose to treat active learning algorithm design as a meta-learning problem and learn the best criterion from data. We model an active learning algorithm as a deep neural network that inputs the base learner state and the unlabelled point set and predicts the best point to annotate next. Training this active query policy network with reinforcement learning, produces the best non-myopic policy for a given dataset. The key challenge in achieving a general solution to AL then becomes that of learner generalisation, particularly across heterogeneous datasets. We propose a multi-task dataset-embedding approach that allows dataset-agnostic active learners to be trained. Our evaluation shows that AL algorithms trained in this way can directly generalise across diverse problems. 
\end{abstract}


\section{Introduction}


In many applications, supervision is costly relative to the data volume.  
Active learning (AL) aims to carefully choose  training data, so that a classifier can perform well even with relatively sparse supervision. This field has collectively proposed numerous  query criteria
, such as margin \citep{Tong:2002:SVM:944790.944793} and uncertainty-based  \citep{kapoor2007algp} sampling, representative and diversity-based \citep{Chattopadhyay:2012:BMA:2339530.2339647} sampling, or combinations thereof \citep{hsu2015active}. It is hard to pick a clear winner, 
because each is based on a reasonable and appealing -- but completely different -- motivation; and there is no method that consistently wins on all datasets. 
%
Rather than hand-designing a criterion, we  take a learning-based approach. We treat active learning method design as a meta-learning problem and train an active learning policy represented by a neural network using deep reinforcement learning (DRL). It is natural to represent AL as a sequential decision making problem since each action (queried point) affects the context (available query points, state of the base learner) successively for the next decision. In this way the active query policy trained by RL can potentially learn a powerful and non-myopic policy. By treating the increasing accuracy of the base learner as the reward, we optimise for the final goal{: the accuracy of a classifier}.
As the class of deep neural network (DNN) models we use includes many classic criteria as special cases, we can expect this approach should be at least as good as existing methods and likely better due to exploiting more information and non-myopic optimisation of the actual evaluation metric. 

The idea of learning the best criterion within a general function class is appealing, and very recent research has had similar inspiration \citep{Bachman17icml}. However crucially it does not provide a  general solution to AL unless the learned criterion generalises across diverse datasets/learning problems. With DRL we can learn an excellent query policy for a given dataset, but this requires the dataset's labels; and if we had those labels we would not need to do AL in the first place. Therefore this paradigm is only useful if a dataset/learner-agnostic criterion can be trained. Thus our research question for AL moves from ``what is a good criterion?'' to ``how to learn a criterion that generalises?''. In this paper we investigate how to train AL query criteria that generalise across tasks/datasets.
Our approach is to define a DNN query criterion that is \emph{paramaterised by a dataset embedding}. By multi-task training our DNN policy on a diverse batch of source datasets, the network learns how to calibrate its strategy according to the statistics of a given dataset. Specifically we adapt the recently proposed auxiliary network idea \citep{Romero17-iclr} to define a meta-network that provides unsupervised domain adaptation. The meta network generates a dataset embedding and produces the weight matricies that parameterise the main policy. This enables an end-to-end query policy to generalize across datasets -- even those with different feature space dimensionality.
Finally, unlike  \citet{DBLPjournals/corr/WoodwardF17} and \citet{Bachman17icml}, our framework is agnostic to the base classifier. Treating the underlying learner as part of the environment to be optimised means that it can be applied to improve the label efficiency of any existing learning architecture or algorithm.

\section{Preliminaries}
\textbf{Reinforcement Learning (RL)}\quad In model-free reinforcement learning, an agent interacts with an environment $\mathcal{E}$ over a number of  time steps $t$. At each  step, it receives the state $s_t\in\mathcal{S}$ from environment and selects an action $a_t\in\mathcal{A}$ based on its policy $\pi(a_t|s_t)$. 
The agent then receives a new state $s_{t+1}$ and reward $r_t$ from $\mathcal{E}$. The aim of RL is to maximise the discounted return $J=\sum_{t=1}^\infty \gamma^{t-1}r_{t}$. 
There are various ways to learn the policy $\pi$. We use direct policy search \citep{kober2009policySearchDMP}, which learns $\pi$ by gradient ascent on an objective function $J_\pi(\theta)$.

\textbf{Active Learning (AL) }\quad A dataset $\mathcal D=\{(\bm{x}_i,y_i)\}_{i=1}^{n}$ contains $n$ instances  $\bm{x}_i\in \mathbbm{R}^D$ and {labels $y_i\in\{1,2,...,C\}$}, most or all of which are unknown in advance. In AL, the data is split between a labelled $\mathcal L$
and unlabelled   
$\mathcal{U}=\mathcal{D}\setminus\mathcal{L}$ set where $|\mathcal{L}|\ll|\mathcal{U}|$ and a classifier $f$
has been trained on $\mathcal{L}$ so far. In each iteration, a pool-based active learner selects an instance/point from unlabelled pool $\mathcal{U}$ to query its label
$(\mathcal{L},\mathcal{U},f)\to i$,
where $i\in \{1,\dots,|\mathcal{U}|\}$.
Then the selected point $i$ is removed from $\mathcal{U}$ and added to $\mathcal{L}$ along with its label, and the classifier $f$ is retrained based on the updated $\mathcal{L}$. 

\textbf{Connection between RL and AL}\quad We  model an AL criterion as a neural network, and discovery of the ideal criterion as an RL problem. Let the  world state $s_t=\{\mathcal{L}_t,\mathcal{U}_t,f_t\}$ contain a featurisation of the dataset and the base classifier. 
An AL criterion is a policy $\pi(a_i|s_t)$ where the discrete action $a_i\in \{1,\dots,|\mathcal{U}_t|\}$ selects a point in $\mathcal{U}_t$ to query. After a query, the state is updated to $s_{t+1}$ as the point is moved from $\mathcal{U}$ to $\mathcal{L}$ and the classifier $f$ updates accordingly.
The reward is  the quantity we wish to maximise, e.g., $Acc_{t}$, the accuracy at query $t$. In this paper we focus on binary classification.
\cut{The reward of an episode is defined as the quantity we wish to maximise. E.g., If the budget is $N$ queries and we care about the accuracy after the $N$th query, then we let $R=Acc_N$ where $Acc_N$ is the accuracy after the $N$th query. 
Alternatively, if we care about the performance during all the $N$ queries, we can use $R=\sum_{t=1}^N  
\gamma^{t-1}Acc_{t}$. (This illustrates an important advantage of the learning to do active learning approach:  we can tune the learned criterion to suit the requirements of the AL application.) } 
\section{Methods}
We aim to train an effective dataset-agnostic active query policy $\pi_\theta(a_t|s_t)$. The key challenge is how to learn a policy $\pi_\theta$ given that: (i) the test and training dataset statistics may differ, and moreover (ii) different datasets have different feature dimensionality $d$. This is addressed by defining the policy $\pi_\theta(a_t|s_t)$ in terms of two sub-networks -- a policy network and meta network.

\begin{figure}[t]
\centering
\includegraphics[width=0.395\columnwidth]{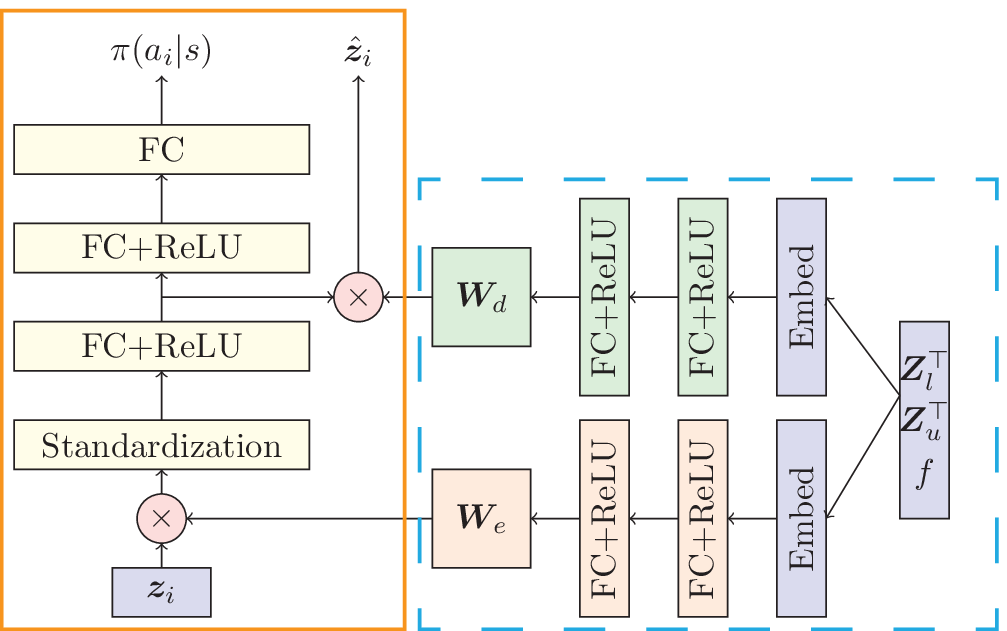}
\includegraphics[width=0.295\textwidth]{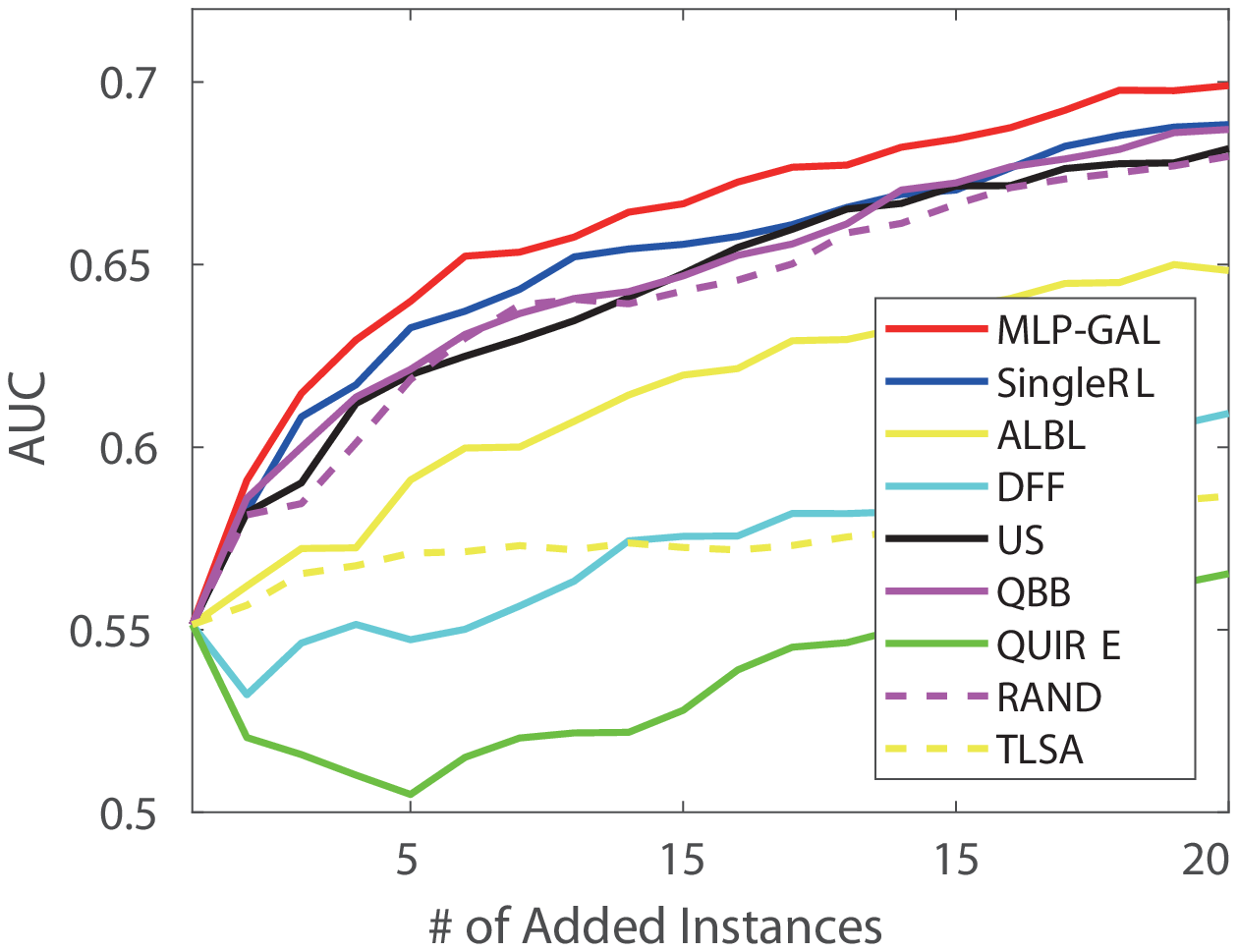}
\includegraphics[width=0.295\textwidth]{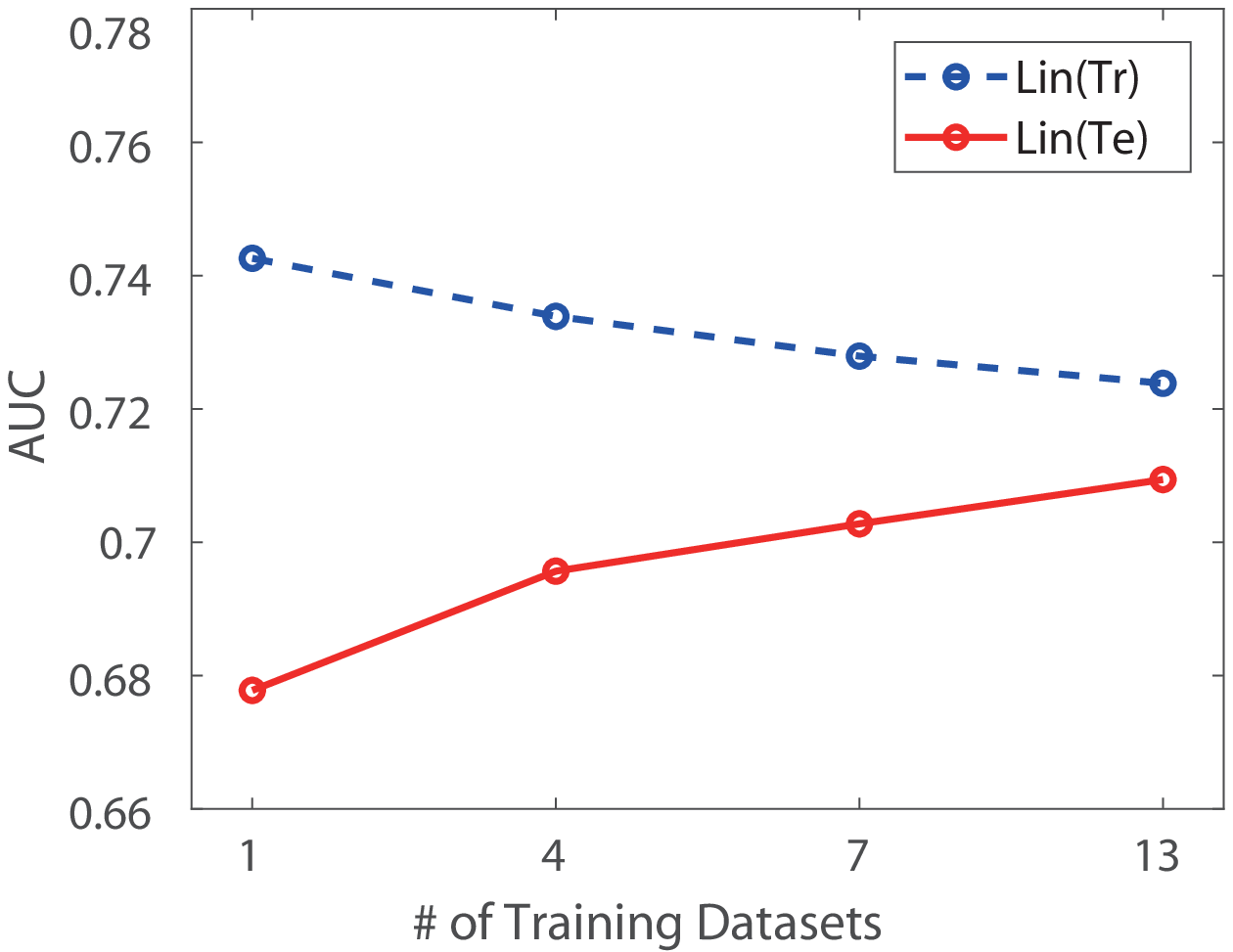}
\caption{Left: Policy and meta-network  for learning a task-agnostic active query policy.  Policy  inputs data-points $z_i$ and outputs a query probability $\pi(a_i|s_t)$. The policy is paramaterised by weights $\bm{W}_e$  generated by the meta network based on the current state $s_t=\{\mathcal{L}_t,\mathcal{U}_t,f_t\}$ which represents the current dataset and classifier. Middle: Illustrative active learning curves from evaluating our learned policy on held out UCI dataset diabetes.
Right: Cross-dataset generalisation. Average performance over all training and testing sets when varying the number of training domains.\label{fig:acc_domain_num0}}
\end{figure}

\keypoint{Policy Network} 
The policy  $\pi$ inputs the $N$ currently unlabelled instances $\bm Z_u\in\mathbbm R^{N\times d}$ and outputs an $N$-way softmax  for  selecting the instance to query.
It selects actions via the softmax  $\pi(a_i|s_t)\propto \exp^{ \Phi_{\theta_p}(\bm W_e^T \bm z_i)} $, where $\bm z_i \in \mathbbm R^d$ is the $i$th unlabelled instance in $\bm Z_u$ and $\bm W_e\in \mathbbm R^{d\times k}$ {are dataset dependent weights.} Although dimensionality $d$ varies by dataset, the encoding  $\bm u_i= \bm W_e^T \bm z_i \in \mathbbm{R}^{k}$ does not, so the rest of the policy network $\pi(a_i|s_t)\propto \exp^{\Phi_{\theta_p}(\bm u_i)}$ is independent of $d$. The key is then how to obtain encoder $\bm{W}_e$ which will be provided by the meta network. 
Following previous work \citep{Bachman17icml,DBLP:journals/corr/KonyushkovaSF17} we also allow the instances to be augmented by instance-level expert features so $\bm Z=[\bm X, \bm \xi(\bm X)]$ where $\bm{X}$ are the raw instances and $\bm \xi(\bm{X})$ are the expert features (distance furthest first and uncertainty) of each raw instance.

\keypoint{Meta Network}  The encoding parameters  $\bm W_e\in\mathbbm{R}^{d\times k}$ of the policy are obtained from the {meta network}: $\Psi_{\theta_m^e}:\{(\mathcal{L}_t,\mathcal{U}_t, f_t) \to \bm W_e ; \theta_m^e\}$. 
Following \citet{Romero17-iclr} we also use the $\bm W_d\in\mathbbm{R}^{k\times d}$ dimensional decoder $\Psi_{\theta_m^d}:\{(\mathcal{L}_t,\mathcal{U}_t, f_t) \to \bm W_d ; \theta_m^d\}$ to regularise this process by reconstructing the input features. {\citet{Romero17-iclr} applied meta networks for parameter reduction, with all training and testing are performed on the same dataset.}  Here the meta network idea is used to \emph{learn how to perform unsupervised deep domain adaptation} across datasets: by synthesising dataset-conditional weight matricies based on dataset-embeddings of $\bm Z^T$ described  next. Note that this deep adaptation is performed in a single feed-forward pass in contrast to existing shallow \citep{csurka2017domainAdaptationBook} or deep but iterative \citep{icml2015_ganin15} methods.

\subsection{Achieving Cross Dataset Generalisation}


\keypoint{Dimension-wise Embedding} The  meta-network  builds a dataset size independent dimension-wise embedding of the input $(\mathcal{L}_t,\mathcal{U}_t, f_t)$, shown in light blue part in Fig.~\ref{fig:acc_domain_num0}. Then it predicts $\bm W_e\in\mathbbm{R}^{d\times k}$
\begin{align}
    (\bm W_e)_j=\Psi\Big([\bm e^1_j(\bm  Z_u^T), \bm e^1_j(\bm  Z_l^T), \bm e^2_j([\bm  Z_u^T, \bm  Z_l^T],f_t)] \Big).
\end{align}
\noindent 
Here $e$ is a non-linear feature embedding,  $j$ indexes features, selecting the $j$th embedded feature and the $j$th row of  $\bm W_e$, {and $\Psi$} is the non-linear mapping of the meta-network, { which outputs a vector of dimension $k$}. Similarly, the meta-network also predicts the weight matrix $\bm W_d$ used for auto-encoding reconstruction 
(Fig~\ref{fig:acc_domain_num0}).
Although $d$ is dataset dependent, the meta network generates weights for a policy network of appropriate dimensionality ($d\times k$) to the target problem. 

\keypoint{Choice of Embeddings}
 We use `representative' and `discriminative' histogram embeddings. 
\emph{Representative} embedding ($\bm e^1_j(\bm  Z_u^T)$ and $ \bm e^1_j(\bm  Z_l^T)$): We encode each feature dimension as a histogram over the instances in that dimension. Specifically, we rescale the $i$th dimension features into $[0,1]$ and divide the dimension into 10 bins. Then we count the proportion of labelled and unlabelled data for each bin. This gives a $1\times 20$ histogram embedding for each dimension {that encodes its moments}.  \emph{Discriminative} embedding ($\bm e^2_j([\bm  Z_u^T, \bm  Z_l^T],f_t)$): We create a 2-D histogram of 10 bins per dimension. In this histogram we count the frequency of instances with feature values within each bin (as per the previous embedding) jointly with the frequency of instances with posterior values within each bin {(i.e., binning on the [0,1] posterior of the binary base classifier.)} 
Finally counts in the $10\times10 $ grid are vectorised to $1\times100$. Concatenating these two embeddings we have 
a $E=120$ dimensional representation of each feature dimension for processing by the meta network.

\textbf{Training for Cross Dataset Generalisation}\quad  We train policy and meta networks jointly using REINFORCE policy search \citep{williams1992reinforce} to maximise the return (active learning accuracy). 
To ensure that our networks achieve the desired learning problem invariance, we perform multi-task training on multiple source datasets: In every mini-batch the return is averaged over a randomly sampled subset of source datasets. Thus \emph{achieving high return means the meta network has learned to synthesise a good per-dataset policy based on the dataset embedding}.\cut{, and that together they generalise across multiple tasks/datasets} 
We further standardise the return from each episode to compensate for diverse return scale across datasets of differing difficulty.
\subsection{Reinforcement Learning Training and Objective Functions}


\keypoint{Reward}
We define the reward the improvement in test split accuracy\cut{ after a query and classifier update}: $r_t=Acc_{t}-Acc_{t-1}$. 
We then optimise the return of an active learning session $J(\theta)=\mathbbm{E}[\sum_{t=1}^\infty \gamma^{t-1}r_t(s,\pi_{\theta}(\cdot,s))]$.

\cut{ The ideal active learner should query the instance that maximally improves the base learner's performance. The reward that reflects the quantity we care about is therefore the increase of test split accuracy $r_t=Acc_{t}-Acc_{t-1}$. To optimise this quantity non-myopically, we define the return of an active learning session as $J(\theta)=\mathbbm{E}[\sum_{t=1}^\infty \gamma^{t-1}r_t(s,\pi_{\theta}(\cdot,s))]$. We then use policy gradient to train the policy and meta-networks to optimise the objective  $J(\theta)$.}

\keypoint{Auxiliary Regularisation Losses} Besides optimising the obtained reward, we also optimise for two auxiliary regularisation losses. \textbf{Reconstruction:} the policy network should reconstruct the unlabelled input data using $\bm{W}_d$ predicted by the meta-network \citep{Romero17-iclr}. {We optimise $A(\bm{Z}_u)=|\bm{Z}_u-\hat{\bm{Z}}_u|_F$, the mean square reconstruction error of the autoencoder.} \textbf{Entropy:} following \citet{pmlr-v48-mniha16}, we also prefer a policy that maintains a high-entropy posterior over actions so as to continue to explore and avoid pre-emptive convergence to an over-confident solution.


With these three objectives, we train both networks where end-to-end, maximising $\theta=\{\theta_p, \theta_m^e\}$ in:\cut{. We maximise the whole objective function $F$ by reversing the sign of reconstruction loss where $\theta=\{\theta_p, \theta_m^e\}$}
\begin{align}
    F = J_\theta(\Phi)
    -\lambda_1 A_{\theta_m^d}(\bm Z_u)+\lambda_2 \mathcal{H}(\pi_\theta(\bm a|\bm Z_u))
    \label{equ:objective func}
\end{align}

\section{Experiments}
\keypoint{Datasets} We experiment with a diverse set of 14 UCI datasets including \emph{austra}, \emph{heart}, \emph{german}, \emph{ILPD}, \emph{ionospheres}, \emph{pima}, \emph{wdbc}, \emph{breast}, \emph{diabetes}, \emph{fertility}, \emph{fourclass}, \emph{habermann}, \emph{livers}, \emph{planning}. {We use  leave-one-out (LOO) setting: training on 13 datasets, and evaluating on the held out dataset.}

\keypoint{Architecture} The auxiliary network for encoder has fully connected (FC) layers of size $120,100,100$ ($E=120,k=100$) and an analogous structure for the decoder.
The policy network has layers of size $N\times d$ ($N\times d$ input matrix $\bm Z_u$),  $N\times100$, $N\times50$, $N\times10$, $N\times1$ ($N$-way output). 
All penultimate layers use ReLU activation. The first FC layer of the policy  is generated by the auxiliary network. Thereafter for efficient implementation with few parameters and to deal with the variable sized input and output, the policy network is implemented convolutionally. We convolve a $h_1\times h_2$ sized filter across the $N$ dimension of each $N\times h_1$ shaped layer to obtain the next $N\times h_2$ layer.


\keypoint{Settings} We use Adam with initial learning rate 0.001 and hyperparameters {$\lambda_1=0.03$,  $\alpha=\lambda_2=0.005$ and discount factor $\gamma=0.99$.} During RL training, we use two tricks to stabilise the policy gradient. 1) We use a relatively large batch size of 32 episodes. 2) We smooth the gradient by accumulation $G_t = (1-\alpha)G_{t-1}+ \alpha g_t$ where $g_t$ is the gradient of the $a_t$ in time step $t$ and the $G_t$ is the accumulated gradient. \cut{Intuitively, the accumulated gradient $G_t$ relies on earlier time step actions.} \textcolor{black}{We train the policy and meta network end-to-end for 50,000 iterations and perform active learning over a time horizon (budget) of $20$. As base learner we use linear SVM 
with class balancing. All results are averages over 100 trials of training and testing dataset splits.} \cut{\textbf{Expert Features:} To enhance the low-level feature of each instance in $\bm{X}$ we define expert features $\xi(\bm{X})$ to include distance furthest first and uncertainty as the augmented feature. }
\subsection{Results}

\keypoint{Alternatives}
We compare with the classic approaches uncertainty/margin-based sampling (US) \citep{Tong:2002:SVM:944790.944793,kapoor2007algp}, furthest-first  (DFF) \citep{Baram:2004:OCA:1005332.1005342} and {query-by-bagging (QBB)  \citep{abe1998query_boostbag}}, as well as random sampling (RAND) as a lower bound. US simply queries the instance with minimum certainty. While simple, it is competitive to more sophisticated criteria and  robust in the sense of hardly ever being a very poor criterion on any benchmark. {We also compare with QUIRE \citep{NIPS2010_4176} as a representative more sophisticated approach, and ALBL \citep{hsu2015active} --- a recent (within-dataset) learning based approach.} 
We denote our method meta-learned policy for general active learning (MLP-GAL). As a related alternative we propose SingleRL. This is our RL approach, but without the meta-network, so a single model is learned over all datasets. Without the meta-network it can only use expert features $\xi(\bm{X})$ so that dimensionality is fixed over datasets. \cut{To give SingleRL an advantage we concatenate some extra global features to the input space\footnote{variance of classifier weight, proportion of labelled pos/neg instances, proportion of predicted unlabelled pos/neg instances, proportion of budget used \citep{DBLP:journals/corr/KonyushkovaSF17}}.} SingleRL can also be seen as a version of one of the few state-of-the-art learning-based alternatives \citep{DBLP:journals/corr/KonyushkovaSF17}, with an important upgrade from supervised learning used there to non-myopic reinforcement learning.

\keypoint{Multi-Task Training Evaluation}
We first verify if it is  possible to learn a \emph{single} policy that generalises across multiple 
training datasets. {In our leave-one-out setting, this means generalising across all  13 datasets in any split.} {Each result in the MLP-GAL~(Tr) column of Table~\ref{tab:SVM} is an average across all the 13 combinations. MLP-GAL learns an effective criterion that outperforms  competitors.} 

\begin{table*}[t]
    \centering
    \caption{Comparison of AL algorithms, leave one dataset out setting. Linear SVM base learner. AUC averages (\%) over 100 trials (and 13 training occurrences for MLP-GAL (Tr)).\cut{\textcolor{blue}{Winner AL algorithm is bolded in each row.}}}\label{tab:SVM}
    \resizebox{\textwidth}{!}{\begin{tabular}{c|c|ccccccccc}
            \hline
	Liinear &MLP-GAL (Tr)	& MLP-GAL (Te)	&        SingleRL (Te)	     &  {Uncertainty}    &    DFF	   &     RAND	     &   ALBL	 & T-LSA    &   QUIRE	      &  QBB	\\         \hline
austra	    & 80.14	& 78.09	& 75.72	& 78.24	& 75.63	& 75.87	& 75.31	& 72.98 & 64.46	& \textbf{78.58}	\\
breast	    & 96.67	& 95.95	& 94.78	& 95.41	& 95.76	& 94.71	& 95.67	& \textbf{96.21} & 95.60	& 95.73	\\
diabetes	& 67.53	& \textbf{65.99}	& 64.78	& 64.18	& 57.31	& 64.05	& 61.35	& 57.34 & 53.75	& 64.46	\\
fertility	& 78.26	& 75.09	& \textbf{77.86}	& 75.79	& 70.44	& 71.28	& 66.92	& 71.18 & 54.93	& 73.87	\\
fourclass	& 74.79	& \textbf{74.11}	& 71.83	& 69.55	& 71.26	& 69.08	& 68.69	& 69.98 & 64.48	& 70.81\\
haberman	& 67.31	& \textbf{65.61}	& 64.91	& 60.16	& 60.26	& 57.40	& 52.49	& 59.67 & 45.89	& 60.58	\\
heart	    & 76.68	& 72.77	& 72.84	& 73.38	& \textbf{73.99}	& 73.06	& 71.78	& 71.52 & 67.07	& 73.36	\\
german	    & 68.01	& \textbf{64.68}	& 63.35	& 63.34	& 61.78	& 62.77	& 61.74	& 58.75 & 51.82	& 64.16	\\
ILPD	    & 62.48	& 59.30	& \textbf{61.08}	& 57.60	& 50.97	& 57.62	& 52.91	& 53.15 & 48.57	& 56.77	\\
ionospheres	& 74.96	& \textbf{71.46}	& 69.78	& 70.47	& 59.64	& 69.81	& 68.44	& 58.95 & 57.84	& 70.40	\\
liver	    & 55.66	& 55.51	& \textbf{55.62}	& 53.45	& 52.87	& 52.87	& 51.25	& 51.36 & 48.11	& 52.13	\\
pima	    & 67.64	& \textbf{67.01}	& 64.67	& 64.18	& 57.31	& 63.69	& 61.27	& 57.03 & 53.75	& 64.24	\\
planning	& 60.74	& \textbf{58.63}	& 56.75	& 55.09	& 52.77	& 54.17	& 49.46	& 52.04 & 39.90	& 55.43	\\
wdbc	    & 90.90	& 90.09	& 88.72	& \textbf{90.93}	& 87.55	& 88.52	& 88.41	& 85.15 & 82.17	& 90.68	\\         \hline
\textbf{Avg}	& 72.98	& \textbf{70.94}	& 70.19 	& 69.41	& 66.25	& 68.21	& 66.12	& 65.38 & 59.17	& 69.37 	\\
\textbf{Num Wins}	&  - 	& \textbf{7}	& 3	& 1	& 1	& 0	& 0	& 1 & 0	& 1	\\
        \hline
    \end{tabular}}
    \label{tab:lin_result}
\end{table*}

\keypoint{Cross-Task Generalisation}
In our leave-one-out setting, each row in Table~\ref{tab:SVM} represents a testing set, and the MLP-GAL~(Te) result is the performance on this test set after training on all 13 other datasets. Our MLP-GAL outperforms alternatives in both average performance and number of wins. {SingleRL is generally also effective compared to prior methods, showing the efficacy of training a policy with RL. However it does not benefit from a meta network, so is not as effective and robust as our MLP-GAL.} 
It is also interesting to see that while sophisticated methods such as QUIRE sometimes perform very well, {they also often perform very badly -- even worse than random. Meanwhile the classic uncertainty and QBB methods perform consistently well.
} This dichotomy illustrates the challenge in building sophisticated AL algorithms that generalise to datasets that they were not engineered on. In contrast, although our MLP-GAL~(Te) has not seen these datasets during training, it performs consistently well due to adapting to each dataset via the meta-network.


\keypoint{Dependence on Number of Training Domains} We next investigate how performance depends on the number of training domains. We train MLP-GAL with an increasing number of source datasets -- 1, 4, 7 (multiple splits each), and (13 splits LOO setting). Then we compute the average performance over all training and all testing domains, in all of their multiple  occurrences across the splits. From the results in Fig~\ref{fig:acc_domain_num0}, we see that the training performance becomes worse when doing a higher-way multi-task training. This is intuitive: it becomes harder to overfit to more datasets simultaneously. Meanwhile testing performance improves, demonstrating that the model learns to generalise better to held out problems when forced to learn on a greater diversity of source datasets.

\section{Discussion}
We have proposed a learning-based perspective on active query criteria design. Our meta-network learns unsupervised domain adaptation: for the first time addressing the key challenge of learning deep query policies with dataset-agnostic generality, rather than requiring dataset-specific training. Our method is also base learner agnostic unlike \citet{Bachman17icml} and \citet{DBLPjournals/corr/WoodwardF17}, so it can be used with any classifier. A limitation thus far (shared by \citep{DBLP:journals/corr/KonyushkovaSF17}) is that we have only focused on a binary base classifier. In the future we would like to evaluate our method on with deep multi-class classifiers as base learners by designing embeddings which can represent the state of such learners, as well as explore application to the stream-based AL setting.



\bibliographystyle{plain}
\bibliography{myref}

\begin{thebibliography}{16}
\providecommand{\natexlab}[1]{#1}
\providecommand{\url}[1]{\texttt{#1}}
\expandafter\ifx\csname urlstyle\endcsname\relax
  \providecommand{\doi}[1]{doi: #1}\else
  \providecommand{\doi}{doi: \begingroup \urlstyle{rm}\Url}\fi

\bibitem[Abe and Mamitsuka(1998)]{abe1998query_boostbag}
Naoki Abe and Hiroshi Mamitsuka.
\newblock Query learning strategies using boosting and bagging.
\newblock In \emph{ICML}, 1998.

\bibitem[Bachman et~al.(2017)Bachman, Sordoni, and Trischler]{Bachman17icml}
Philip Bachman, Alessandro Sordoni, and Adam Trischler.
\newblock Learning algorithms for active learning.
\newblock \emph{ICML}, 2017.

\bibitem[Baram et~al.(2003)Baram, El-Yaniv, and
  Luz]{Baram:2004:OCA:1005332.1005342}
Yoram Baram, Ran El-Yaniv, and Kobi Luz.
\newblock Online choice of active learning algorithms.
\newblock \emph{Journal of Machine Learning Research}, 5:\penalty0 255--291,
  2003.

\bibitem[Chattopadhyay et~al.(2012)Chattopadhyay, Wang, Fan, Davidson,
  Panchanathan, and Ye]{Chattopadhyay:2012:BMA:2339530.2339647}
Rita Chattopadhyay, Zheng Wang, Wei Fan, Ian Davidson, Sethuraman Panchanathan,
  and Jieping Ye.
\newblock Batch mode active sampling based on marginal probability distribution
  matching.
\newblock KDD. ACM, 2012.

\bibitem[Csurka(2017)]{csurka2017domainAdaptationBook}
Gabriela Csurka.
\newblock \emph{Domain Adaptation in Computer Vision Applications}.
\newblock Springer, 2017.

\bibitem[Ganin and Lempitsky(2015)]{icml2015_ganin15}
Yaroslav Ganin and Victor Lempitsky.
\newblock Unsupervised domain adaptation by backpropagation.
\newblock In \emph{ICML}, 2015.

\bibitem[Hsu and Lin(2015)]{hsu2015active}
Wei-Ning Hsu and Hsuan-Tien Lin.
\newblock Active learning by learning.
\newblock In \emph{AAAI}, 2015.

\bibitem[Huang et~al.(2010)Huang, Jin, and Zhou]{NIPS2010_4176}
Sheng-jun Huang, Rong Jin, and Zhi-hua Zhou.
\newblock Active learning by querying informative and representative examples.
\newblock In \emph{NIPS}. 2010.

\bibitem[Kapoor et~al.(2007)Kapoor, Grauman, Urtasun, and
  Darrell]{kapoor2007algp}
Ashish Kapoor, Kristen Grauman, Raquel Urtasun, and Trevor Darrell.
\newblock Active learning with gaussian processes for object categorization.
\newblock In \emph{ICCV}, 2007.

\bibitem[Kober and Peters(2009)]{kober2009policySearchDMP}
Jens Kober and Jan~R. Peters.
\newblock Policy search for motor primitives in robotics.
\newblock In \emph{NIPS}. 2009.

\bibitem[Konyushkova et~al.(2017)Konyushkova, Sznitman, and
  Fua]{DBLP:journals/corr/KonyushkovaSF17}
Ksenia Konyushkova, Raphael Sznitman, and Pascal Fua.
\newblock Learning active learning from real and synthetic data.
\newblock \emph{NIPS}, 2017.

\bibitem[Mnih et~al.(2016)Mnih, Badia, Mirza, Graves, Lillicrap, Harley,
  Silver, and Kavukcuoglu]{pmlr-v48-mniha16}
Volodymyr Mnih, Adria~Puigdomenech Badia, Mehdi Mirza, Alex Graves, Timothy
  Lillicrap, Tim Harley, David Silver, and Koray Kavukcuoglu.
\newblock Asynchronous methods for deep reinforcement learning.
\newblock In \emph{ICML}, 2016.

\bibitem[Romero et~al.(2017)Romero, Carrier, Erraqabi, Sylvain, Auvolat,
  Dejoie, Legault, Dube, Hussin, and Bengio]{Romero17-iclr}
Adriana Romero, Pierre~Luc Carrier, Akram Erraqabi, Tristan Sylvain, Alex
  Auvolat, Etienne Dejoie, Marc{-}Andr{\'{e}} Legault, Marie{-}Pierre Dube,
  Julie~G. Hussin, and Yoshua Bengio.
\newblock Diet networks: Thin parameters for fat genomics.
\newblock \emph{ICLR}, 2017.

\bibitem[Tong and Koller(2002)]{Tong:2002:SVM:944790.944793}
Simon Tong and Daphne Koller.
\newblock Support vector machine active learning with applications to text
  classification.
\newblock \emph{J. Mach. Learn. Res.}, 2, March 2002.

\bibitem[Williams(1992)]{williams1992reinforce}
Ronald~J Williams.
\newblock Simple statistical gradient-following algorithms for connectionist
  reinforcement learning.
\newblock \emph{Machine Learning}, 1992.

\bibitem[Woodward and Finn(2017)]{DBLPjournals/corr/WoodwardF17}
Mark Woodward and Chelsea Finn.
\newblock Active one-shot learning.
\newblock \emph{CoRR}, abs/1702.06559, 2017.

\end{thebibliography}

\end{document}